\documentclass[letterpaper]{article} 
\usepackage{aaai2026}  
\usepackage{times}  
\usepackage{helvet}  
\usepackage{courier}  
\usepackage[hyphens]{url}  
\usepackage{graphicx} 
\usepackage{natbib}  
\usepackage{caption} 
\usepackage{algorithm}
\usepackage{algorithmic}

\usepackage{newfloat}
\usepackage{listings}
\DeclareCaptionStyle{ruled}{labelfont=normalfont,labelsep=colon,strut=off} 
\floatstyle{ruled}
\newfloat{listing}{tb}{lst}{}
\floatname{listing}{Listing}
\title{ConsistTalk: Intensity Controllable Temporally Consistent Talking Head Generation with Diffusion Noise Search}
\author {
    Zhenjie Liu,
    Jianzhang Lu,
    Renjie Lu,
    Cong Liang,
    Shangfei Wang\thanks{The corresponding author.}
}
\affiliations {
    University of Science and Technology of China\\
    \{liuzhenjie, jzlu, lu\_h, lc150303\}@mail.ustc.edu.cn, sfwang@ustc.edu.cn
}

\usepackage{bibentry}

\usepackage{amsmath}
\usepackage{amssymb}
\usepackage{multirow}
\usepackage{booktabs}

\begin{document}

\maketitle

\begin{abstract}
Recent advancements in video diffusion models have significantly enhanced audio-driven portrait animation. However, current methods still suffer from flickering, identity drift, and poor audio-visual synchronization. These issues primarily stem from entangled appearance-motion representations and unstable inference strategies. In this paper, we introduce \textbf{ConsistTalk}, a novel intensity-controllable and temporally consistent talking head generation framework with diffusion noise search inference. First, we propose \textbf{an optical flow-guided temporal module (OFT)} that decouples motion features from static appearance by leveraging facial optical flow, thereby reducing visual flicker and improving temporal consistency. Second, we present an \textbf{Audio-to-Intensity (A2I) model} obtained through multimodal teacher-student knowledge distillation. By transforming audio and facial velocity features into a frame-wise intensity sequence, the A2I model enables joint modeling of audio and visual motion, resulting in more natural dynamics. This further enables fine-grained, frame-wise control of motion dynamics while maintaining tight audio-visual synchronization. Third, we introduce a \textbf{diffusion noise initialization strategy (IC-Init)}. By enforcing explicit constraints on background coherence and motion continuity during inference-time noise search, we achieve better identity preservation and refine motion dynamics compared to the current autoregressive strategy. Extensive experiments demonstrate that ConsistTalk significantly outperforms prior methods in reducing flicker, preserving identity, and delivering temporally stable, high-fidelity talking head videos.
\end{abstract}


\section{Introduction}
Audio-driven portrait animation aims to synthesize expressive talking head videos from a static portrait image and accompanying speech audio. Early works \cite{sadtalker,aniportrait} primarily employ a two-stage pipeline using 3D motion or keypoints as intermediate representations. However, representational bottlenecks limit the ability of these methods to produce dynamic and realistic facial movements.

Recent advancements in this field have been spurred by the emergence of image generation diffusion models such as Stable Diffusion (SD) \cite{ldm} and DiT \cite{dit}. Diffusion models have demonstrated superior generative capabilities compared to non-diffusion models, leading to their adaptation for portrait animation tasks. For example, EMO \cite{emo}, the first end-to-end framework capable of generating high-realism animations from a single image and audio, extends SD to video generation by incorporating temporal modules. Likewise, other notable approaches \cite{echomimic, hallo, loopy, sonic, fantasytalking} built upon the pretrained image diffusion model backbone demonstrate robust generation capabilities in this domain.

\begin{figure}[!t]
  \centering
  \includegraphics[width=\linewidth]{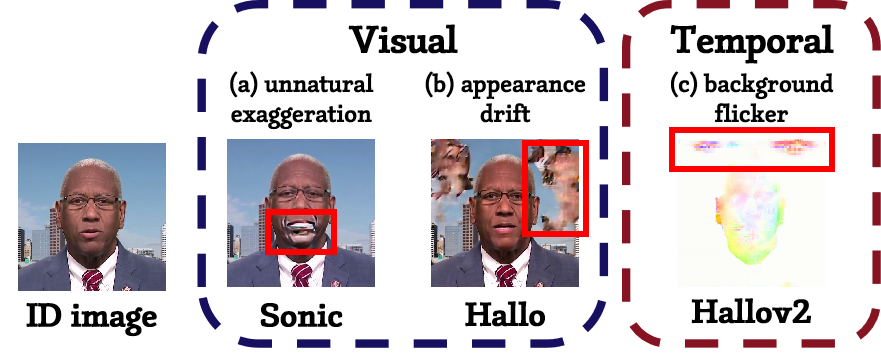}
    \caption{Failure cases in maintaining visual and temporal consistency of current methods. The left samples exhibit significant visual distortion (a) and background deformation with identity drift (b). The rightmost visualization displays temporal flickering (c) in the optical flows.}
    \label{fig:motivation}
\end{figure}

Despite these advancements, existing methods still exhibit significant temporal and visual inconsistencies, as illustrated in Figure \ref{fig:motivation}. First, predominant approaches leverage motion features extracted from preceding frames through the ReferenceNet as temporal signals. However, flickering remains problematic due to the contamination of appearance information in complex scenes \cite{hallo2}. Second, current methods either predict motion solely from audio or from previous frames, resulting in abrupt temporal jitter. Controlling motion intensity may disrupt the synchronization between audio and visual elements, leading to unstable inferences. Finally, many methods \cite{hallo,hallo2,fantasytalking} adopt an autoregressive-like inference strategy for long video generation to maintain consistent motion across segments. By recursively using generated frames as reference images, deformations will accumulate over time and cause noticeable appearance drift.

To address these challenges, we propose \textbf{ConsistTalk}, a novel intensity-controllable and temporally consistent talking head generation framework with a diffusion noise search inference strategy. The method comprises three key innovations: 1) \textbf{Optical Flow-guided Temporal Module (OFT)}: We utilize facial optical flow to decouple facial dynamics from static appearance features, thereby preventing appearance contamination in the temporal layer and reducing flicker. 2) \textbf{Audio-to-Intensity (A2I) Module}: Intensity modeling captures the one-to-many mapping between audio and facial movements. A student model is distilled from a multimodal teacher to predict a frame-wise intensity sequence from audio alone, enabling controllable motion while maintaining audio-visual synchronization. 3) \textbf{IC-Init strategy}: We introduce a training-free inference algorithm that leverages a noise searching mechanism guided by the intensity sequence. By enforcing constraints on background stability and motion continuity, we enhance generation robustness and dynamic quality. Extensive experiments demonstrate that ConsistTalk outperforms existing methods in temporal consistency, identity preservation, and motion fidelity. Our contributions are summarized as follows.
\begin{itemize}
\item We present \textbf{ConsistTalk}, a novel diffusion-based model that leverages facial optical flow to decouple motion from appearance, thereby mitigating temporal flickering. Through multimodal learning, it effectively captures the correlation between audio and motion velocity and enables motion dynamics control.

\item We propose \textbf{IC-Init}, a training-free diffusion noise initialization strategy that leverages an intensity-guided noise search, significantly improving identity preservation and motion continuity during inference.

\item We conduct fair and extensive experiments across diverse settings to validate our approach. Results show that ConsistTalk consistently generates more vivid, expressive, and temporally coherent talking head videos than existing state-of-the-art methods.
\end{itemize}

\begin{figure*}[!t]
  \centering
  \includegraphics[width=0.95\textwidth]{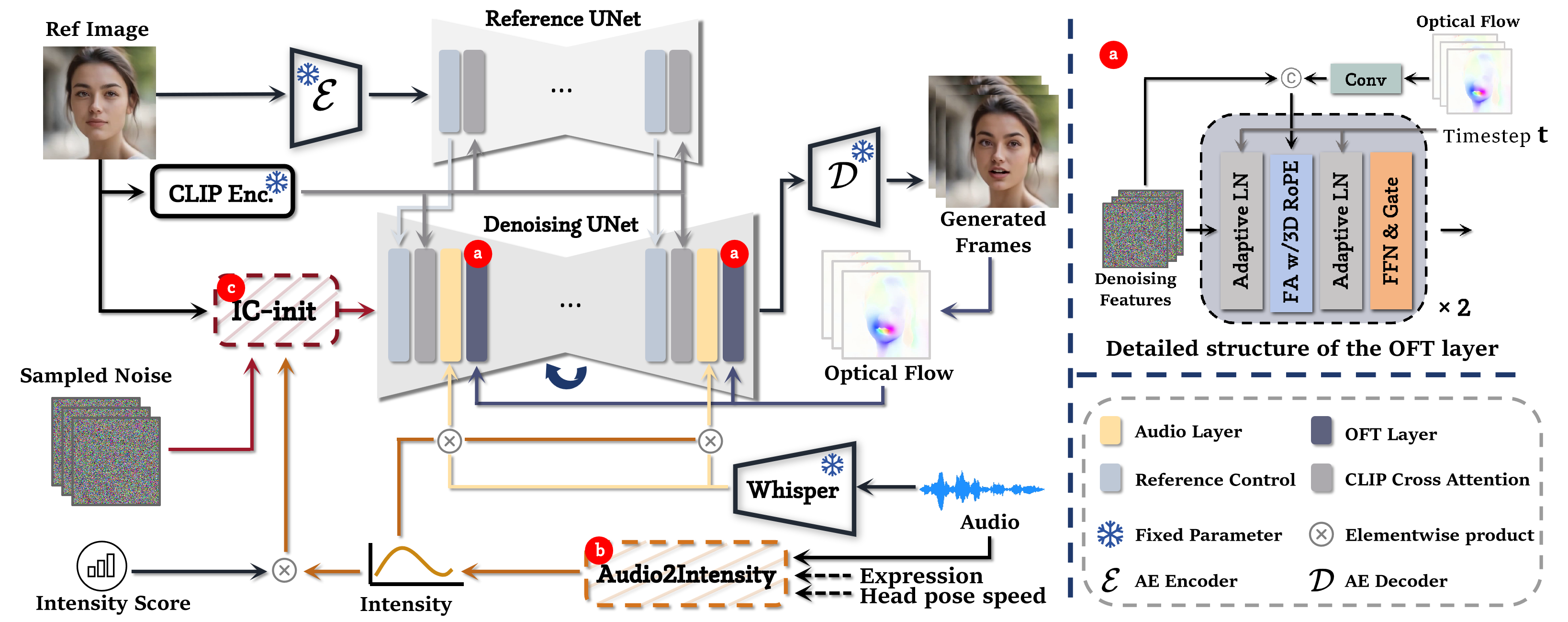}
    \caption{Overall architecture of the proposed ConsistTalk framework. The system integrates three key modules: (a) a facial optical flow-guided temporal module (OFT) that decouples motion dynamics from appearance features to suppress flicker and enhance temporal consistency; (b) a audio-to-intensity (A2I) model that transforms audio and facial velocity features into frame-wise intensity signals for fine-grained motion control; and (c) a noise initialization strategy (IC-Init) that stabilizes the generation process by leveraging a noise search procedure.}
    \label{fig:main}
\end{figure*}

\section{Related Work}

Audio-driven talking head generation has seen significant progress in recent years. Existing approaches generally fall into two categories based on their generation strategies: two-stage pipelines and end-to-end generation frameworks.

\textbf{Two-Stage Talking Head Generation.} Early works like SadTalker \cite{sadtalker} and Aniprotrait \cite{aniportrait} in this field primarily incorporated intermediate 3D motion representations and employed a two-stage pipeline for video generation. Specifically, in the first stage, audio features are converted into motion parameters, while in the second stage, these motion cues guide a rendering or warping-based synthesis module to generate video frames. However, the reliance on intermediate motion representations limits their ability to model nuanced facial dynamics.

\textbf{End-To-End Talking Head Generation.} Recent studies \cite{emo,hallo,hallo2,loopy,sonic,letstalk} have integrated diffusion models into this field. By leveraging the powerful generative capabilities of pretrained image diffusion models, these approaches employ Stable Diffusion \cite{ldm} or Wan 2.1 \cite{wan} as the backbone network. To enhance the image generation model's ability to capture cross-frame dynamics, these methods incorporate ReferenceNet features from previous frames to construct temporal layers. However, because they rely on overall visual features as motion control signals without decoupling motion features, they still suffer from temporal inconsistencies. Figure \ref{fig:motivation} (c) illustrates cross-frame optical flow, showing sudden variations in the background that result in noticeable temporal flickering. In contrast, our method explicitly separates motion dynamics from appearance via facial optical flow, improving background coherence. This fundamentally differs from implicit decoupling strategies such as Patch-Drop in Hallo2 \cite{hallo2}.

Secondly, current methods often struggle to achieve visually consistent control of motion intensity. Hallo \cite{hallo} proposes a hierarchical audio-driven visual synthesis module, while Sonic \cite{sonic} introduces a motion-decoupled control module that predicts motion buckets through audio-reference image fusion. However, as illustrated in Figure \ref{fig:motivation} (a), unnatural or exaggerated motions can occur due to unstable motion intensity control strategies. In contrast, we present a multimodal teacher-student knowledge distillation learning strategy designed to explicitly predict motion intensity. The acquired intensity value is then provided to the denoising model as a latent scale factor, offering a user-friendly method for controlling motion intensity while maintaining audio-visual synchronization.

Thirdly, for long-form video generation, autoregressive-like inference \cite{hallo,hallo2} is commonly used, where previously generated frames are recursively fed as motion conditions or reference images. Although this approach facilitates continuous motion across frames, it is susceptible to error accumulation. Deformations relative to the reference image or inconsistencies in motion from earlier frames can propagate to subsequent frames, leading to issues with background and identity shifting, as shown in Figure \ref{fig:motivation} (b). We enhance this inference strategy by constructing a noise search algorithm that incorporates explicit constraints related to the reference image and preceding frames, thereby improving motion diversity and temporal consistency.

\section{Methodology}
In this section, we present ConsistTalk for audio-driven portrait image animation, designed to mitigate flickering and temporal inconsistency problems effectively. Let $\boldsymbol{I}_{ref}$ denote the reference image, and let $\{\boldsymbol{I}_{t-1},\cdots,\boldsymbol{I}_{t-n}\}$ represent the preceding $n$ generated frames. Our architecture is built upon the Dual-UNet design \cite{emo,hallo,echomimic}. We first use CLIP \cite{clip} to extract features from the reference ID portrait image. Compared with Hallo\cite{hallo}'s face embedding, it is compatible with the stable diffusion's text-image cross-attention space. To decouple temporal motion dynamics from identity information in motion frames, we introduce a facial optical flow-guided temporal module (Sec. \ref{sec3:of}). In addition, a multimodal joint modeling module is proposed to learn audio-video intensity from both audio signals and facial/head motion sequences (Sec. \ref{sec3:intensity}). The intensity value is then used to guide noise initialization, enhancing the quality of generation and ensuring temporal consistency (Sec. \ref{sec3:search}).

\subsection{Optical Flow Guided Temporal Module}\label{sec3:of}

The temporal module in current methods builds upon ReferenceNet, an extension of ControlNet \cite{controlnet}, which is widely used in image-to-video (I2V) generation to enable spatio-temporal alignment \cite{animate_anyone,magicanimate,improving,emo}. To avoid appearance contamination in spatio-temporal fusion, we adopt a decoupling strategy: identity (ID) features are encoded by ReferenceNet only \textbf{\textit{once}}, while motion information is dynamically constructed by estimating facial optical flow between adjacent frames. Optical flow captures inter-frame sub-pixel-level motion dynamics information from pixel space. The core idea is to decouple motion dynamics from identity information in the generated frames, thus ensuring that the model relies primarily on the reference image for appearance features and utilizes preceding frames to capture temporal movement features.

Specifically, the global visual texture information is extracted with the help of ReferenceNet and injected into the Dual-UNet through cross-attention. Then we use a lightweight model FacialFlownet \cite{facialflownet} to estimate the facial optical flow frames $\{\boldsymbol{F}_i\in\mathbb R^{C\times H\times W}\}_{i=t-1}^{t-n}$ and downsample them to the resolution of the latent space using 3D convolution. To enhance temporal consistency, we employ the Full-Attention mechanism \cite{hunyuanvideo, opensoraplan,cogvideox} with 3D RoPE \cite{rope}. Compared to the divided spatiotemporal (2D+1D) attention, it can better integrate spatial and temporal features and improve the temporal consistency of the generated video. Refer to Figure \ref{fig:main} (a) for a visualization of the temporal module.

\begin{figure*}[!tp]
  \centering
  \includegraphics[width=0.95\textwidth]{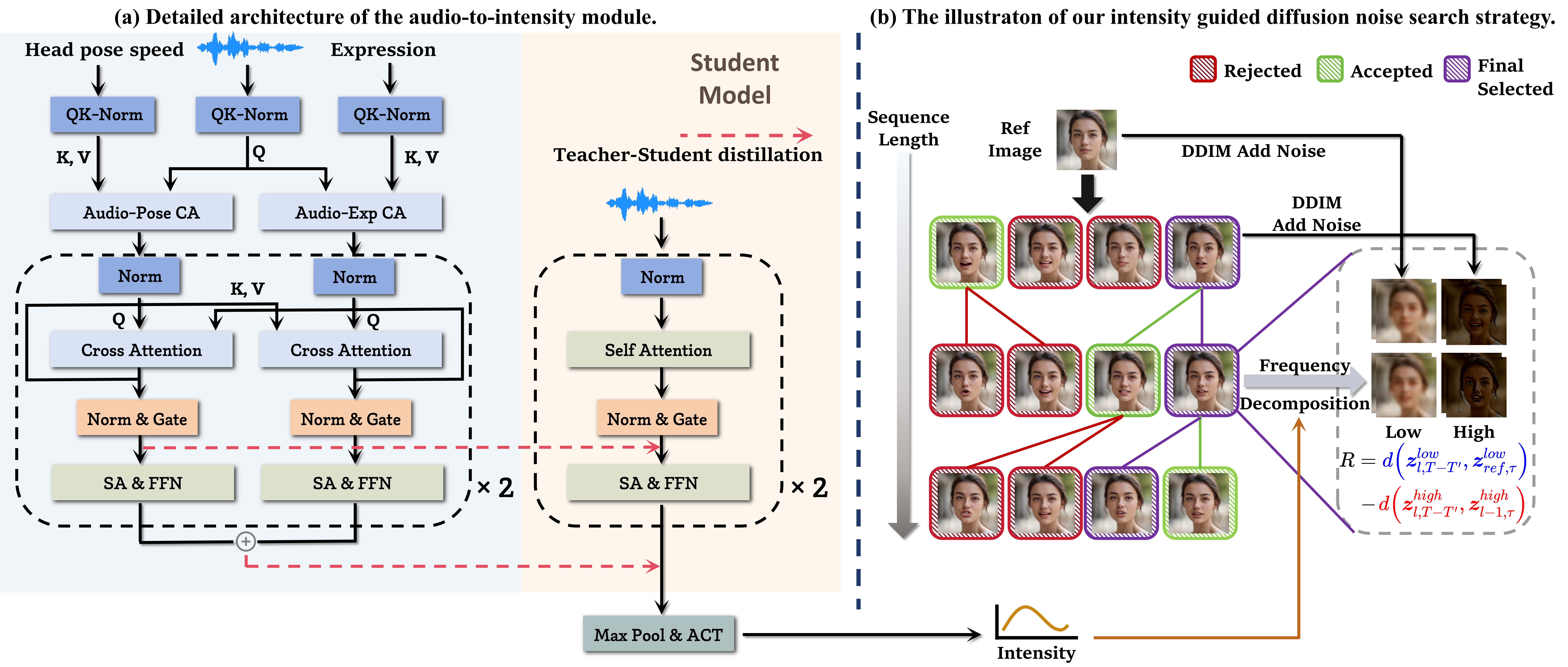}
    \caption{Audio-to-Intensity module and IC-Init. The left side presents the detailed structure of the audio-to-intensity module, and on the right is the inference-time noise search strategy IC-Init based on intensity-guided frequency decomposition.}
      \label{fig:a2i}
\end{figure*}

\subsection{Motion Intensity Sequence Modeling}\label{sec3:intensity}
Here, we propose a motion intensity sequence modeling strategy, in which the intensity sequence captures the overall magnitude of head and facial expression movements over time. Unlike prior methods \cite{sonic,loopy,hallo}, our approach explicitly estimates a frame-wise intensity sequence and performs multimodal fusion during training, rather than relying on modal dropout as in \cite{loopy}. Specifically, we first use Whisper-Tiny \cite{whisper} to extract audio features, and a duration of $0.2s$ windowed audio features is used for each frame and projected using linear layers. The transformed audio embedding is denoted as $c_a\in\mathbb R^{b\times f\times d_a}$, where $b$ denotes batch size, $f$ is segment length, and $c$ is cross attention dimension. Additionally, following Loopy \cite{loopy}, facial velocity-related features, such as head motion velocity $v_h\in\mathbb R^{b\times L\times 1}$ and facial landmark motion variance $v_f\in \mathbb R^{b\times L\times 1}$, are extracted, where $L$ denotes the velocity feature length. Then we use 1D convolution and linear layers to match the shape of the projected audio features.

As shown in Figure \ref{fig:a2i}, we propose an audio-to-intensity module based on teacher-student learning. The teacher model leverages multimodal cross attention and a variant of the gated linear unit to capture the relationships between driven audio and portrait movement-related features. We use the Swish \cite{swish} activation function and take $\boldsymbol{y}=\texttt{Swish}(\boldsymbol{h})$ as the output intensity sequence. Due to the audio-velocity multimodal fusion, intensity can be interpreted as an aggregate representation of the overall magnitude of head pose dynamics and facial expression movements. Since the intensity sequence has no ground truth as the supervisory signal, we draw inspiration from time series tasks \cite{dilate} to make the intensity sequence and the input velocity sequence construct shape and temporal loss. Specifically, the intensity loss is defined by
\begin{align}
\mathcal L_{intensity}=&\alpha\sum\nolimits _{c\in\{h,f\}}\mathcal L_{shape}(\boldsymbol y,v_c) \notag \\
&+ (1-\alpha)\sum\nolimits_{c\in\{h,f\}}\mathcal L_{temporal}(\boldsymbol y,v_c)
\end{align}
where $\boldsymbol y$ is the output intensity sequence, and $c$ represents the velocity modal. Since velocity-related features are unavailable during the inference phase, we employ a knowledge distillation approach to train a student model that predicts the intensity sequence solely from audio input. Specifically, we minimize the mean squared error (MSE) loss between the \textbf{cross-attention} logits of the last $l$ layers of the teacher and student models, as demonstrated in Figure \ref{fig:a2i} (a). This design encourages the student model to learn the multi-modal correspondence and anchors the learned intensity to concrete velocity priors. We further enforce that the intensity sequence matches the velocity patterns via shape and temporal losses. Formally, the total loss function for the student model is defined as $\mathcal L=\mathcal L_{MSE}+\alpha\cdot \mathcal L_{intensity}$.

\begin{algorithm}[!ht]
\caption{Intensity Guided Diffusion Noise Beam Search}
\label{alg:icinit}
\begin{algorithmic}[1]
\REQUIRE latent denoising diffusion model $\boldsymbol \epsilon_{\theta}$, reference portrait image latent $\boldsymbol{z}_{ref}$, number of beams $B$, number of candidates $K$, number of lookahead steps $T'$, number of add noise steps $\tau$

\STATE $\boldsymbol z_{ref,\tau}=\texttt{add\_noise}(\boldsymbol z_{ref},\tau)$
\STATE $B_0=\{\boldsymbol z_{0}^s \}_{s=1}^B$ {\;\;$\triangleright$ Initial Beams} 

\FOR{$l=1$ {\bfseries to} $L$} 
    \STATE $\texttt{budget}=\varnothing,\; B_l=\varnothing$
    \FOR{$\boldsymbol z_{l-1}^s \in B_{l-1}$}
        \STATE $\boldsymbol z_{l-1,\tau}^s=\texttt{add\_noise}(\boldsymbol z_{l-1}^s,\tau)$
        
        {$\triangleright$ Sample $K$ candidates for each latent in $B_{l-1}$} 
        \STATE $\big\{\boldsymbol z_{l,T}^{s,k} \sim \mathcal N(\boldsymbol 0, \boldsymbol I)\big\}_{k=1}^K$ 

        {$\triangleright$ lookahead denoising} 
        \STATE $\boldsymbol z_{l,T-T'}^{s,k}=\texttt{de\_noise}\left(\boldsymbol z_{l,T}^k,\boldsymbol\epsilon_{\theta},T'\right)$ 
        \STATE $\texttt{budget}\; += \Big\{(\boldsymbol z_{l,T-T'}^{s,k}, \boldsymbol z_{l-1,\tau}^s)\Big\}_{k=1}^K$
    \ENDFOR
    
    {$\triangleright$ Search $B$ higher reward from $B\times K$ pairs}
    
    \FOR{ $\boldsymbol z_{l,T-T'}^{s,k}, \boldsymbol z_{l-1,\tau}^s \in \texttt{top-}B\Big(\texttt{budget}, R(\cdot)\Big)$}
        \STATE $B_l\; += \texttt{de\_noise}\left(\boldsymbol z_{l,T-T'}^{s,k},\boldsymbol \epsilon_{\theta},T-T'\right)$
    \ENDFOR
\ENDFOR

{$\triangleright$ Final selected path}
\STATE {\bfseries return: } $\{\boldsymbol z'_{l} \}_{l=1}^L=\texttt{argmax}_{\{z_{l}\}_{l=1}^L\subset \bigcup B_l } \sum R(\cdot)$

\end{algorithmic}
\end{algorithm}

\begin{table*}[h]
\begin{center}
\begin{tabular}{lc|cc|cc|c|cc}
\toprule[1.5pt]
\multirow{2}{*}{\textbf{Method}} & \multicolumn{1}{c|}{Lip Sync}                & \multicolumn{2}{c|}{Quality}   & \multicolumn{2}{c|}{Consistency}               & \multicolumn{1}{c|}{ID}               & \multicolumn{2}{c}{Motion}       \\
\cline{2-9}
\specialrule{0em}{1pt}{1pt}
& \textbf{Sync-C}$\uparrow$ & \textbf{FVD}$\downarrow$ & \textbf{E-FID}$\downarrow$ & \textbf{Flicker}$\downarrow$ & \textbf{VBench} $\uparrow$& \textbf{ID-dist}$\downarrow$ &\textbf{BA}$\uparrow$ & \textbf{Div}$\uparrow$ \\

\midrule

SadTalker$^{\dagger}$ \shortcite{sadtalker}& 5.44& 778.2& 2.029& \underline{0.5445}& 99.5 / 97.77& 0.2204& 1.468 & \underline{3.31}\\  
AniTalker$^{\dagger}$ \shortcite{anitalker}& 4.53& 575.0& 2.569& 0.7519& 99.5 / \textbf{98.37}& 0.2021& 1.432& 2.75 \\   
Hallo \shortcite{hallo}& 6.3& 268.4&2.327& 2.3274& 99.37 / 97.18& 0.1864 & 1.545 & 2.59 \\   
Hallo-v2 \shortcite{hallo2}& 6.27& 284.1& 2.5685& 0.5914& 99.4 / 96.95& 0.1889 & 1.452 & 3.01 \\   
EchoMimic \shortcite{echomimic}& 6.12&693.8& \textbf{1.138}&0.6552& 99.34 / 97.13& \underline{0.1777} & 1.507 & 2.95 \\   
Sonic \shortcite{sonic}& \textbf{7.11}&\underline{206.1}& 1.664&0.688& 99.61 / 97.47& 0.1835 & 1.41 & 2.8 \\  
FantasyTalking \shortcite{fantasytalking}& 5.82&256.9& 2.069&1.874& \underline{99.63} / 96.56& 0.228 & \underline{1.581} & 2.43 \\  
\hline
\specialrule{0em}{1pt}{1pt}
\textbf{ConsistTalk (Full Model)} & \underline{6.83}& \textbf{171.7} & \underline{1.4397}&\textbf{0.4218}& \textbf{99.68} / \underline{98.28}& \textbf{0.1743}& \textbf{1.659} & \textbf{3.48} \\   
Base Model +OFT +A2I& 6.92&186.5& 1.3758&0.6583& 99.64 / 97.68& 0.1723& 1.624 & 3.16 \\  
Base Model +OFT& 5.94&203.9& 1.3542&0.5288& 99.67 / 97.44& 0.1705& 1.551 & 2.95 \\   
Base Model (w/ReferenctNet)& 6.28&228.4& 1.4637&2.2642& 99.26 / 97.15& 0.1823 & 1.533 & 2.81 \\     
                                
\bottomrule[1.5pt]
\end{tabular}
\caption{Quantitative comparison results with baselines on HDTF \cite{hdtf} dataset. The best result for each metric is in \textbf{bold}, and the second-best is \underline{underlined}. The \texttt{Base Model} refers to the original Dual-Unet framework from Hallo, without optical flow motion feature decoupling (i.e., the standard ReferenceNet strategy). And \texttt{Full Model} represents the base model with all proposed modules (OFT, A2I, and IC-Init). \hfill $^{\dagger}$: evaluated using $256\times256$ resolution outputs without resizing.} \label{tab:quantitative}
\end{center}
\end{table*}

The acquired intensity value is provided to the denoising UNet model by multiplying it with the hidden states of the audio-conditioned layer. This design enables explicit, frame-wise control over motion dynamics and diversity by adjusting the intensity values during inference.

\subsection{Intensity Guided Diffusion Noise Search}\label{sec3:search}

In this subsection, we present our noise initialization strategy, \textbf{IC-Init}. Drawing inspiration from ConsistI2V \cite{consisti2v}, we observe that high-frequency components encode rapid motion and fine details, while low-frequency components preserve coarse structure and identity. Based on this observation, we perform frequency decomposition and define a reward function conditioned on the reference image and previous frames. The reward weights are adaptively modulated by the predicted intensity sequence. This design enables intensity-aware beam search to select temporally coherent and visually dynamic latent paths during inference. 

The procedure is illustrated in Figure \ref{fig:a2i} (b) and Algorithm \ref{alg:icinit}. Specifically, we establish the diffusion noise beam search process in the temporal dimension. At each sequence step $l = 1, \ldots, L$, we sample $K$ latent candidates from a Gaussian distribution for every beam in the previous set (Line 7). These candidates are partially denoised with a $T'$-step lookahead (where $T' \ll T$) to form a search budget of size $B \times K$ (Lines 8-9). Additionally, we apply $\tau$-step noise to both the reference latent and the latent from the previous timestep (Lines 1 and 6). From the budget, we then select the top-$B$ candidates based on a reward function:
\begin{eqnarray}
&R(\boldsymbol z_{l,T-T'},\boldsymbol z_{ref,\tau},\boldsymbol z_{l-1,\tau})  =\notag\\
&\lambda_{low}\cdot\left\langle\boldsymbol z_{l,T-T'}^{low},\boldsymbol z_{ref,\tau}^{low}\right\rangle -\lambda_{high}\cdot\left\langle \boldsymbol z_{l,T-T'}^{high},\boldsymbol z_{l-1,\tau}^{high}  \right\rangle
\end{eqnarray}

The high- and low-frequency components are obtained via:
\begin{eqnarray}
\boldsymbol z =\texttt{FFT\_3D}(\boldsymbol z)\odot \mathcal M,\; \mathcal M=\begin{cases}
\mathcal G(y_l),&\text{low\;freq.}\\
1-\mathcal G(y_l),&\text{high\;freq.}
\end{cases}
\end{eqnarray}
where $\mathcal G(y):\mathbb R^1\to\mathbb \{0,1\}^{b\times c\times f\times h\times w}$ is the Gaussian low-pass filter parameterized by the acquired intensity sequence, and $\langle\cdot,\cdot\rangle$ represents the dot product operation. The reward function comprises two components: a low-frequency similarity term between the reference image latent and the candidate, which preserves identity and global structure, and a high-frequency dissimilarity term that penalizes deviations from the prior frame, thereby promoting motion expressiveness. Maximizing this reward balances stability with dynamic motion over time. The reward weights $\lambda_{low}(y_l)$ and $\lambda_{high}(y_l)$ are dynamically computed based on the normalized intensity score: $\lambda_{low}(y) =  \texttt{sigmoid}\Big(\text{z-norm}(\boldsymbol{y})\Big),\; \lambda_{high}(y) = 1-\lambda_{low}(y)$, where $\text{z-norm}(y) = \frac{y - \mu(y)}{\sigma(y)}$ is the Z-score normalization. Higher intensity emphasizes high-frequency motion, while lower intensity prioritizes low-frequency stability. Finally, we select the path with the highest cumulative reward as the final sequence of latents (Lines 15-16). Compared to the greedy search, IC-Init explores a broader latent space and achieves superior temporal stability and motion fidelity through frequency-aware sampling.

Let $\Omega$ denote the FLOPs of a single forward pass of the diffusion model, and $T$ represents the total denoising steps. Compared to autoregressive inference with complexity $\mathcal{O}(L \cdot T \cdot \Omega)$, IC-Init introduces a beam search with total complexity $\mathcal{O}(L \cdot B \cdot (K \cdot T' + T - T') \cdot \Omega)$, where $B$ and $K$ are beam and candidate sizes, and $T'$ denotes the lookahead steps. Although computational cost approximately increases by a factor of $B$, IC-Init yields significantly better temporal stability, motion diversity, and identity preservation.

\begin{figure}[!ht]
  \centering
  \includegraphics[width=\linewidth]{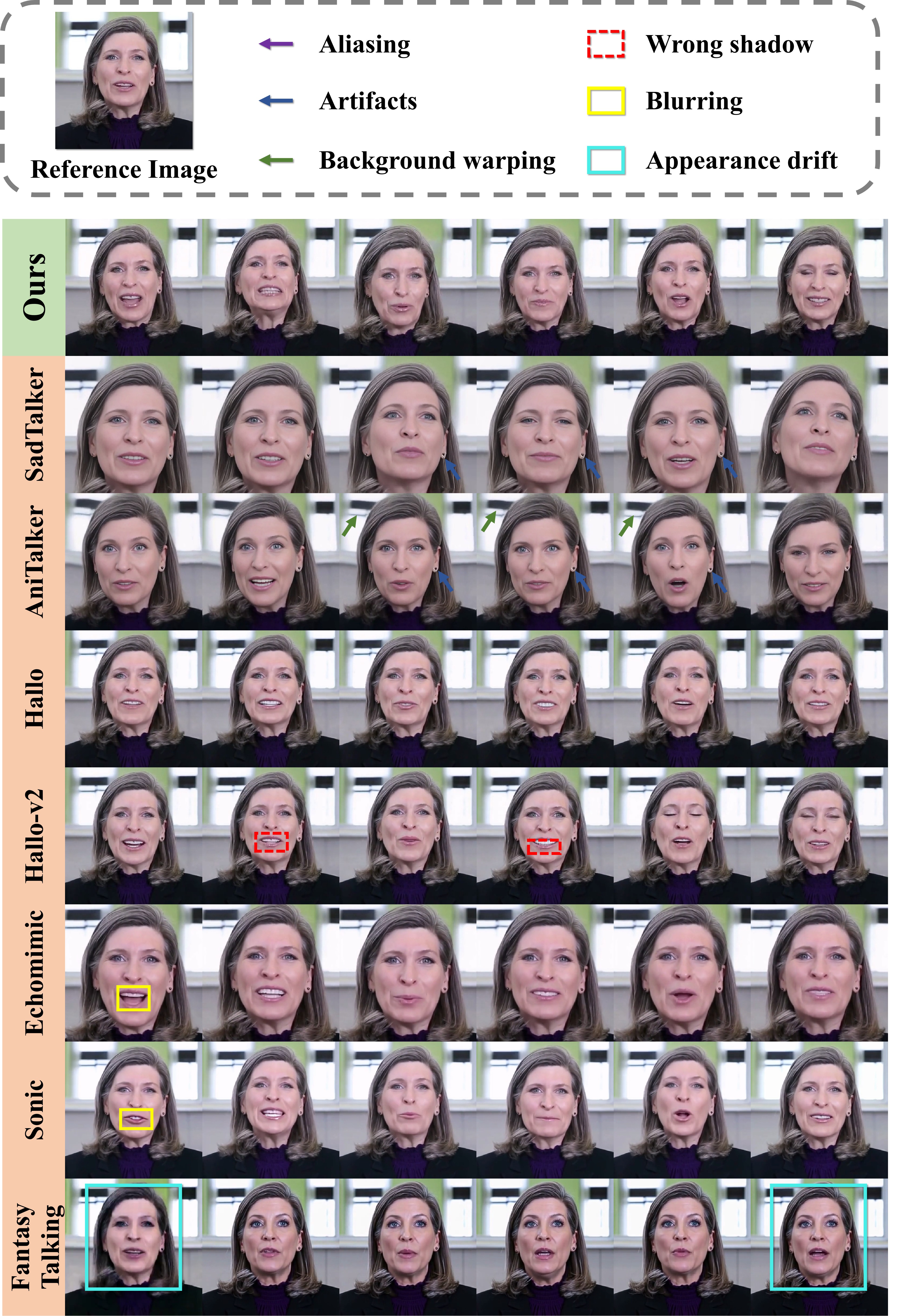}
    \caption{Qualitative comparisons with State-of-the-Art talking head generation methods on HDTF \cite{hdtf} dataset.}
    \label{fig:comparison}
\end{figure}

\section{Experiment}
\subsection{Experimental Setups}
\subsubsection{Datasets}
We developed a diverse training dataset by aggregating data from various open-source datasets, including \textbf{CelebV-HQ} \cite{celebvhq}, \textbf{VFHQ} \cite{vfhq}, and the \textbf{Hallov3} \cite{hallo3} training set. To ensure a fair evaluation, we randomly select 50 16-second clips from \textbf{HDTF} \cite{hdtf} as the test set, ensuring that no identity or video clip is used during training. Following FOMM \cite{fomm}, we first convert each video to 25 FPS and resample the audio to 16 kHz. Then, we crop and resize the facial region to $512\times 512$ resolution using FaceAlignment \cite{face_alignment}. 

\subsubsection{Implementation Details}
Our denoising net is initialized using Hallo-v1. We manipulate the data in both training phases such that the drop ratio of the driven audio, reference image, and optical flows is set to 5\%, respectively. And following Loopy \cite{loopy}, we use DWPose \cite{dwpose} to detect facial keypoints. The velocity of the absolute position of the nose tip across the last 64 frames is the head pose movement feature. The velocity of the displacement of the upper half of the facial keypoints is used as the expression feature. We trained our model using the Adam \cite{adam} optimizer with a learning rate of $1e-5$.

For IC-Init, we use DDIM \cite{ddim} scheduler with $\tau=20,\; T'=3,\; T=30$. And the beam search hyperparameters are $K=4,B=2$.

During training, each instance generates $f=10$ video frame segments and uses ground-truth optical flow from the previous six frames. While in the inference stage, optical flow is computed from previously generated frames.

\subsubsection{Evaluation}
We utilize nine metrics to evaluate. For overall video quality, we measure Fr\'{e}chet Video Distance (FVD) \cite{fvd}, which quantifies the similarity between generated and ground-truth videos, and lower values indicate better realism. We measure Expression FID (E-FID) \cite{emo} and ArcFace \cite{arcface} identity cosine distance (ID-Dist) for facial expression and identity preservation, respectively. SyncNet-C \cite{syncnet} evaluates audio-lip synchronization in generated videos. For evaluating video motion fluidity and temporal consistency, we compute VBench \cite{vbench}'s smooth and background metrics, along with Flicker metrics. In addition, we report Beat Align Score (BA) \cite{bailando} and a motion diversity metric to assess alignment between audio and head motion, and to capture the range of motion dynamics, respectively.


We compare our method against state-of-the-art audio-driven talking head approaches with publicly available implementations: SadTalker \cite{sadtalker}, AniTalker \cite{anitalker}, Hallo(v1\&v2) \cite{hallo,hallo2}, EchoMimic \cite{echomimic}, Sonic \cite{sonic} and FantasyTalking \cite{fantasytalking}.

\begin{figure*}[!t]
  \centering
  \includegraphics[width=\textwidth]{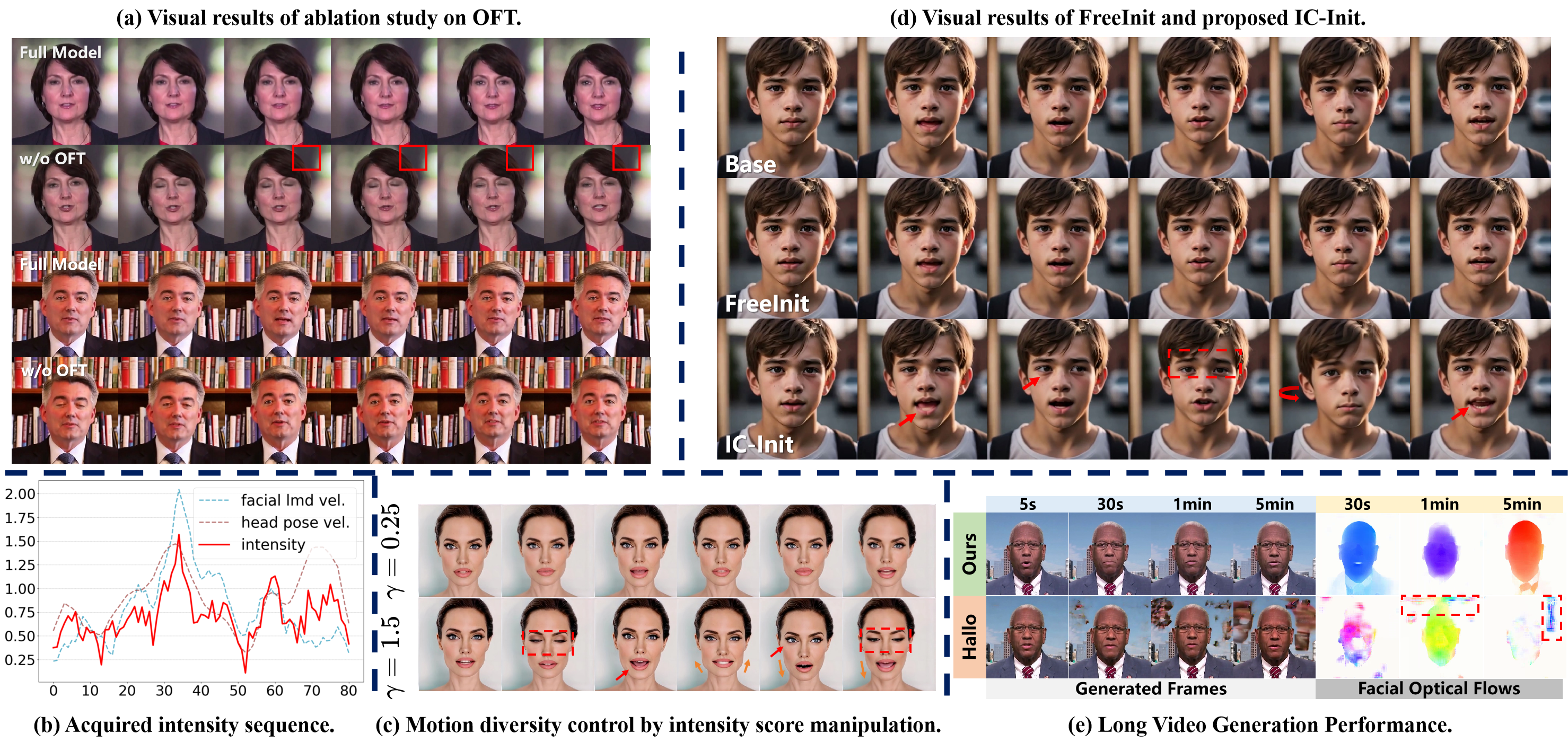}
    \caption{Ablation studies. (a): Visual results for ablation on facial optical-guided temporal module OFT. (b): Line chart of the acquired intensity sequence. (c): Qualitative comparison with intensity score manipulation. (d): Qualitative comparison with FreeInit and proposed IC-Init. (e): Comparison of long video generation quality between ConsistTalk and baseline Hallo.}
    \label{fig:ablation}
\end{figure*}

\subsection{Results and Analysis}
\textbf{Quantitative Comparison. }
We show the quantitative comparison results in Table \ref{tab:quantitative}. ConsistTalk outperforms or matches state-of-the-art methods across multiple metrics. Regarding overall video quality, our framework achieves the lowest FVD score of 171.7, representing a 16.7\% reduction compared to the second-best model, Sonic (206.1), demonstrating superior generation quality. Our method also achieves the lowest ID-Dist score, underscoring its strength in identity preservation and validating the effectiveness of our IC-Init strategy. For temporal stability, as measured by Flicker (0.4218), our model reduces flicker by over 22\% compared to the next best, SadTalker, and by nearly 39\% compared to Sonic. Additionally, VBench scores of 99.68 and 98.28 further confirm superior cross-frame background steadiness. Lip synchronization quality remains high, with a Sync-C score of 6.83, approximately 13.2\% higher than the baseline model Hallo (6.3), indicating that the A2I module preserves accurate mouth movements. Moreover, the model achieves the highest Beat Align (1.659) and Diversity (3.48) scores, reflecting more varied and better-synchronized head motions with audio.

\textbf{Qualitative Comparison. }
Qualitative results are shown in Figure \ref{fig:comparison}. Compared to existing methods, our approach achieves significantly better visual consistency. Prior methods exhibit various artifacts, including aliasing and blurring (e.g., EchoMimic, Sonic), background warping (e.g., AniTalker, SadTalker), identity inconsistency (e.g., FantasyTalking), and lip or shading distortions (e.g., Hallo-v2). In contrast, our method mitigates these issues through three key components: the OFT module decouples motion from appearance to reduce flicker, the A2I module enables synchronized and controllable motion from audio, and IC-Init enhances stability via intensity-guided noise search. These modules collectively ensure natural expressions, stable identity, and high motion fidelity across frames.

\subsection{Ablation Studies}
\textbf{Effectiveness of Optical Flow Guided Temporal Module.} The ablation study in Table \ref{tab:quantitative} quantitatively evaluates the effectiveness of our proposed temporal module, OFT. Compared to the baseline model with ReferenceNet (FVD: 228.4, Flicker: 2.2642), incorporating OFT alone significantly reduces flickering (Flicker: 0.5288, a 76.6\% reduction) and improves generation quality (FVD: 203.9, a 10.7\% reduction). These results demonstrate that explicitly modeling motion via facial optical flow effectively decouples temporal dynamics from appearance features, leading to enhanced visual stability and temporal coherence. The instances in Figure \ref{fig:ablation} (a) demonstrate that OFT mitigates flickering and enhances temporal consistency.

\textbf{Analyses of Audio-to-Intensity Module.} Quantitative results in Table~\ref{tab:quantitative} further validate the contribution of the A2I module. When added to the base model with OFT, A2I improves the Beat Align (BA) score from 1.551 to 1.624, reflecting enhanced synchronization between audio and head motion. Additionally, it increases motion diversity (from 2.95 to 3.16) without degrading identity or temporal consistency, demonstrating that A2I enables fine-grained, frame-wise motion control while preserving natural dynamics.

Figure \ref{fig:ablation} (b) presents a line chart visualization of the acquired intensity sequence along with the corresponding velocity feature sequence. Both the intensity and velocity features exhibit a strong correlation in terms of shape and timing, which supports the validity of intensity loss $\mathcal L_{intensity}$. Figure \ref{fig:ablation} (c) illustrates the qualitative outcomes of motion diversity control achieved by manipulating intensity scores. It reveals that higher intensities significantly enhance the range of motion and overall dynamism, whereas lower intensities restrict movement primarily to the mouth and lip regions.

\textbf{Effectiveness of IC-Init.} We compare three variants of ConsistTalk to evaluate the effectiveness of IC-Init: (1) the base model without IC-Init, (2) a variant with FreeInit using $\tau=20$ and $D_0=0.25$, and (3) the full model with IC-Init. As shown in Figure \ref{fig:ablation} (d), IC-Init stabilizes the output video, minimizes sudden changes in appearance and motion, and produces videos with richer dynamics compared to FreeInit. 

\textbf{Long Video Generation Performance.} We further evaluate the robustness of our method in long video generation scenarios. As shown in Figure \ref{fig:ablation} (e), Hallo suffers from severe appearance drift and background flickering over time, especially after 1–5 minutes. This degradation stems from cumulative errors in its autoregressive inference, leading to identity shifts and distorted frames.

In contrast, ConsistTalk maintains stable identity and coherent facial structure throughout. The corresponding optical flow maps show that our method produces concentrated, face-focused motion fields, while Hallo exhibits noisy and dispersed flows, indicating unstable dynamics. These results highlight the effectiveness of our OFT and IC-Init modules in preserving temporal consistency and suppressing drift in long video generation.
\section{Conclusion}

In this paper, we present \textbf{ConsistTalk}, a novel intensity-controllable and temporally consistent talking head generation framework with diffusion noise search algorithm. Our method integrates three key components: an Optical Flow-guided Temporal module (OFT) for decoupling motion and appearance, a multimodal Audio-to-Intensity (A2I) module for synchronized and controllable motion generation, and a training-free IC-Init strategy for stable inference through intensity-guided noise search. Together, these designs address common challenges such as flickering, background instability, and motion inconsistency. Extensive experiments demonstrate that ConsistTalk produces temporally coherent, identity-preserving, and expressive talking head videos, with fine-grained motion control and superior generation quality.

\section*{Acknowledgements}
This work was supported by National Natural Science Foundation of China 62376255.

\bibliography{aaai2026.bib}
\end{document}